\documentclass[a4paper,conference]{IEEEtran}

\usepackage{subfigure}
\usepackage{graphicx}
\usepackage{CJKutf8}  
\usepackage{pinyin}
\usepackage{enumerate}
\usepackage{bm}
\usepackage{graphicx}
\usepackage{multirow}

\usepackage{soul}
\usepackage{url}
\usepackage[hidelinks]{hyperref}
\usepackage[utf8]{inputenc}
\usepackage[small]{caption}
\usepackage{graphicx}
\usepackage{amsmath}
\usepackage{amsthm}
\usepackage{booktabs}
\usepackage{algorithm}
\usepackage{algorithmic}
\usepackage{color}
\usepackage{cite}

\ifCLASSINFOpdf

\else

\fi

\begin{document}
\IEEEoverridecommandlockouts

\title{CKG: Dynamic Representation Based on Context and Knowledge Graph}


\author{\IEEEauthorblockN{Xunzhu Tang\IEEEauthorrefmark{1},
Tiezhu Sun\IEEEauthorrefmark{2},
Rujie Zhu\IEEEauthorrefmark{3} and
Shi Wang\IEEEauthorrefmark{4}}
\IEEEauthorblockA{\IEEEauthorrefmark{1} Huazhong University \\of Science and Technology\\Wuhan, China\\ Email: realdanieltang@gmail.com}
\IEEEauthorblockA{\IEEEauthorrefmark{2} Momenta, Suzhou, China\\
Email: tiezhu.sun@uni.lu}
\IEEEauthorblockA{\IEEEauthorrefmark{3}University of Central Florida, Orlando, FL, USA\\ Email: rujie.zhu@ucf.edu}
\IEEEauthorblockA{\IEEEauthorrefmark{4}Institute of Computing Technology, Chinese Academy, Beijing, China \\Email:wangshi@ict.ac.cn}}


\maketitle

\begin{abstract}
Recently, neural language representation models pre-trained on large corpus can capture rich co-occurrence information and be fine-tuned in downstream tasks to improve the performance. As a result, they have achieved state-of-the-art results in a large range of language tasks. However, there exists other valuable semantic information such as similar, opposite, or other possible meanings in external knowledge graphs (KGs). We argue that entities in KGs could be used to enhance the correct semantic meaning of language sentences. In this paper, we propose a new method CKG: Dynamic Representation Based on \textbf{C}ontext and \textbf{K}nowledge \textbf{G}raph. On the one side, CKG can extract rich semantic information of large corpus. On the other side, it can make full use of inside information such as co-occurrence in large corpus and outside information such as similar entities in KGs. We conduct extensive experiments on a wide range of tasks, including QQP, MRPC, SST-5, SQuAD, CoNLL 2003, and SNLI. The experiment results show that CKG achieves SOTA 89.2 on SQuAD compared with SAN (84.4), ELMo (85.8), and BERT$_{Base}$ (88.5).
\end{abstract}


%
\IEEEpeerreviewmaketitle

\section{Introduction}

The language representation models, such as ELMo \cite{Peters:2018}, OpenAI-GPT \cite{radford2018improving} and BERT \cite{devlinetal2019bert} have achieved good performance on a large range of NLP tasks including sentence classification, question answering, and sentence tagging. Existing language models mainly focused on pre-training on unstructured corpora with limited available information. For example, polysemy is a pervasive phenomenon. To mitigate it, current models require pre-training embedding on large corpus and then fine-tune them on downstream tasks. It is time-consuming and also difficult to capture more senses of a word by pre-training and fine-tuning \cite{Peters:2018,devlinetal2019bert}. For instance, word apple has different meanings, including a fruit, a brand or IT products. The corpora for training embedding should be collected carefully to make sure different kinds of senses of a word in question are included. After pre-training, it still may be unclear how to interpret the embedding in downstream tasks. Considering sentence ``People in cities usually buy apples and other fruits in local markets'', downstream tasks need much cost to fine-tune it before they learn the exact meaning of word `apple'. Therefore, performance improvement may be hindered by using unstructured training corpus only.

Existing methods predict the masked word based only on its left or/and right context which are local and can not predict the exact meaning of the word in the question. It will restrict the power of the trained embedding if local context is not large enough. The major limitation is that the context from few surrounded words is local. Now, a wealth of knowledge is available in different forms ({\it i.e.,} wikipedia, Freebase) on the web due to the advent of open data initiative. And it is growing rapidly. For example, the volume of data connected by Linked Data \footnote{http://linkeddata.org/home}, a practice for sharing and connecting data using RDF specification \footnote{https://www.w3.org/2001/sw/wiki/RDF}, increases from 93 datasets (March 2009) to 1,239 datasets (as of March 2019). It is thus imperative to advance pre-training.

We propose model CKG to refine semantic of words by knowledge graph, which can make use of global knowledge. The proposed language models are not pre-trained on unstructured texts but on refined ones using knowledge graph, which can strengthen the meaning of each polysemous word in its context. We use BiLSTM to pre-train our model for reason that it uses both left-to-right and right-to-left LSTMs to generate features, and long input will worsen the pre-training performance. Naturally, we introduce 1D-CNN layer before BiLSTM for extracting features so that long input is shortened. We compare CKG with ELMo, BERT, and other models on wide variety of tasks. CKG achieves significant performance improvement compared to existing methods.

The contributions of our work are as follows:
\begin{itemize}
    \item We propose CKG to mitigate polysemy by extending the words using knowledge graph and then fusing the semantic of words. 
    \item We optimize the training process by introducing 1D-CNN layer before BiLSTM to extract features so that dimension reduction can reduce computational costs further.
    \item We compare our method with ELMo, BERT, word2vec, GloVe in multiple tasks ({\it i.e., }QQP, MRPC, and SST-5). Further, CKG+ELMo get SOTA 89.2 on SQuAD task compared with SAN(84.4), pure ELMo(85.8), and $BERT_{BASE}$(88.5).
\end{itemize}

\section{Related Work}
Word representation has been the most important foundation for NLP tasks. Context-independent representations \cite{mikolov2013distributed,pennington2014glove} provide pre-trained word vectors from unlabeled text. They achieved good performance on several major NLP benchmarks including SNLI and SQuAD. These approaches have been generalized to sentence embedding and even paragraph embedding. However, these approaches for learning word vectors from large corpus only allow one context-independent representation for each word, making it difficult to understand text containing polysemous words. This problem is called unsupervised word sense disambiguation.

Some previous works have focused on enriching embedding with subword information \cite{bojanowski2017enriching}, learning separate vectors for each word sense \cite{neelakantan2015efficient} and splitting word into several word senses based on graph clustering \cite{chang2018efficient}. In addition, \cite{annervaz2018learning} uses world knowledge in the form of KGs to enhance learning models. Furthermore, \cite{zhangetal2019ernie, sun2019ernie} employs general KGs information to improve BERT. Inspired by \cite{chang2018efficient,annervaz2018learning,zhangetal2019ernie}, our approach extends senses of central words by traversing its closest neighbors  in knowledge graph. Such an approach can enrich embedding with different senses.

Other recent works have paid attention on learning context-dependent representations. Sentence or document encoders have been pre-trained on unlabeled texts and fine-tuned for a supervised downstream task to generate contextual token representations \cite{radford2018improving}. The advantage of these models is that few parameters need to be learned from scratch and they can mitigate polysemy problem to some extent. For example, ELMo \cite{Peters:2018} uses two BiLSTM layers to encode the context from large corpus on character level for semantic scalability, and fine-tunes word embedding on downstream tasks. Therefore, the model do not only captures complex characteristics of a word use but also mitigates polysemous problems by its dynamic word representations in different contexts. ELMo is a feature-based approach, which means that the pre-trained word representations can be added to downstream models as additional features. Different from ELMo, BERT \cite{devlinetal2019bert} takes bidirectional encoder representations from transformers, which is designed to pre-train on large unlabeled corpus with masked language model on both left and right context in all layers. BERT created state-of-the-art performance in a large range of NLP tasks, but it costs too much time to do pre-training and fine-tuning. And similar to ELMo, when it comes to complex text with polysemous words, these approaches become inefficient.

In this paper, we will take full advantage BiLSTM for text encoding and benefit from knowledge graph to mitigate polysemous problems on NLP tasks. We call the approach in this paper as CKG. CKG is divided into two main parts, of which one is used to extend and refine semantic of the polysemous entities in a given text and the other extracts feature for dimension before pre-training. These two parts will be introduced as follows, and at same time we will show that CKG work well across a broad range of diverse NLP tasks.

\section{Details of CKG Model}
In this section, we introduce CKG and its detailed implementation, including the model architecture in Section \ref{sec:modelarc}, the semantic extension designed for capturing more information from KGs in Section \ref{sec:extension} and semantic fusion designed for extracting the correct information out of the extensions in Section \ref{sec:fusing}, and the details of the pre-training procedure in Section \ref{sec:pre-training}.

\begin{figure}[ht!]
    \centering
    \includegraphics[scale=0.4]{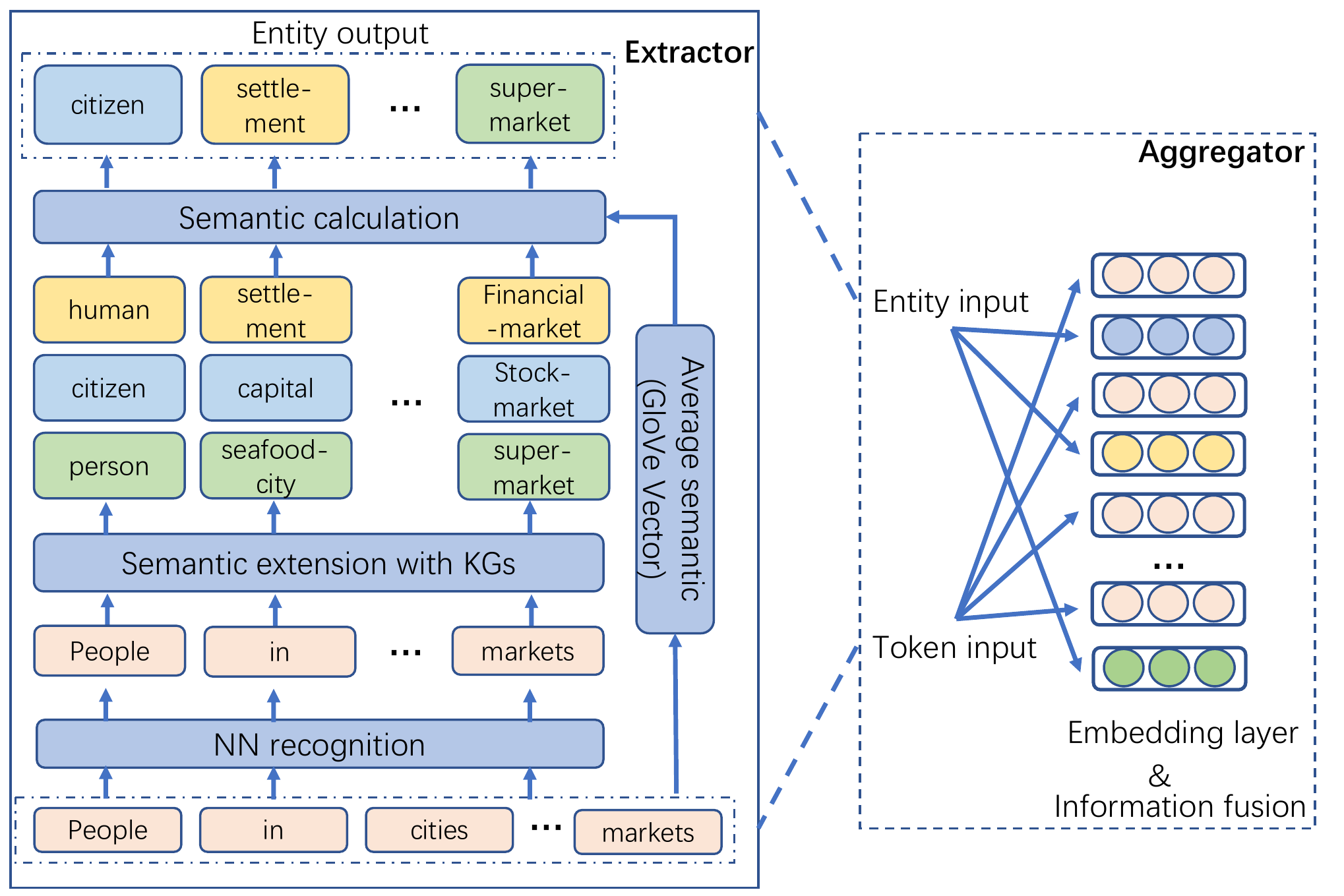}
    \caption{Architecture of CKG, which consists of two main parts ({\it i.e.,} Extractor and Aggregator).}
    \label{fig:model1}
\end{figure}

\subsection{Model Architecture} \label{sec:modelarc}
As shown in Figure \ref{fig:model1}, the whole architecture of CKG consists of two main modules: (1) the semantic extension (\textbf{Extractor}) responsible to capture multiple concrete meanings of some entities they could have, and (2) semantic fusion (\textbf{Aggregator}) responsible to figure out the most possible ones of those extended meanings in a concrete sentence or context and fusing the results with central entities in footnote style. 

To make the description more specific, we denote the input sequence as \{$w_1$,$w_1$,...,$w_n$\} \footnote{$w_i$ means the \textit{i}-th token and in this paper, tokens are at the word level.}, the tokens of extended entities as \{$e_1,e_2,...,e_m$\} \footnote{We take the name style from \cite{zhangetal2019ernie}}. We employ two kinds of initial ways of embedding  to represent the words \footnote{In this paper, we use nouns to represent the central entities to be extended.} in corpus and their extended entities. One is GloVe vector responsible to do calculation in Aggregator, and another one is embedding function from Pytorch responsible to initialize fused input tokens before pre-training. Given a token sequence \{$w_1$,$w_1$,...,$w_n$\}, we figure out its corresponding entity sequence \{$e_{1,1},e_{1,2},e_{1,3},e_{2,2},...,e_{m,1},e_{m,2},e_{m,3}$\} by \textbf{Extractor}. 

By similarity calculation, we finally get only one extension for each central entity. Input tokens and extended entities can get features \{\bm{$w_1,w_2,...,w_n$}\}, \{\bm{$e_{1,1},...,e_{m,3}$}\} as follows:

\vspace{-0.3cm}
\begin{small}
\begin{align}
& \{\bm{w_1,w_1,...,w_n}\} = GloVe(\{w_1,w_2,...,w_n\})  \label{con:glovew} \\
& \{\bm{e_{1,1},...,e_{m,3}}\} = GloVe(\{e_{1,1},...,e_{m,3}\}) \label{con:glovee}
\end{align}
\end{small}where \textit{GloVe}\textit{(}$\cdot$\textit{)} is a initial word-vector function which can map a word to GloVe vector \cite{pennington2014glove}, and $e_{m,1},e_{m,2},e_{m,3}$ \footnote{We only choose the top 3 extensions of each central entity.} are top 3 extensions of central entity m.

Firstly, CKG sums GloVe vectors of $w_1, w_2,...,w_n$ and figure the cosine value between each of \{$e_{m,1}, e_{m,2},e_{m,3}$\} ($e_{m,i}$ is the {\it i}-th hop extension of entity {\it m}) and the average vector of $w_1, w_2,...,w_n$. Then, it employ the one with minimum angle to represent the meaning of entity \textit{m} in sequence \{$w_1$, $w_2$,...,$w_n$\}.

After computing \{$e_{1,1},...,e_{m,3}$\} with average vector of \{$w_1$,$w_2$,...,$w_n$\}, CKG gets the most suitable extensions to represent the origin
entities in the given sequence. Aggregator fuses these extensions with origin entities in the given sequence to inject the meanings of extensions out of KGs. We can get fused sequence as follows,

\vspace{-0.3cm}
\begin{small}
\begin{align}
& Aggregator(\{w_1,w_2,...,w_n\},\{e_{1,1},...,e_{m,3}\}) \notag \\
& =\{\bm{w_1,e_{1,2},...,w_n,e_{m,3}}\} \label{con:newseq} 
\end{align}
\end{small}where \textit{Aggregator}\textit{(}$\cdot$\textit{)} is a function responsible for fusing extensions with their corresponding central entities. More details of \textbf{Extractor} and \textbf{Aggregator} will be introduced in Section \ref{sec:extension} and \ref{sec:fusing}.

\subsection{Semantic Extension Using Knowledge Graph (Extractor)} \label{sec:extension}
To understand the true meaning or sense of a word with multi meanings in a sequence, we should know potential meanings of the word. Both context-independent \cite{mikolov2013distributed,pennington2014glove} and contextual \cite{Peters:2018,radford2018improving,devlinetal2019bert} representation models can not capture all the meanings of entities in a sequence. Therefore, we extend meanings of the entities in a sequence through KGs as more as possible \footnote{In this work, the one of KGs we take is Dbpedia.}. However, it remains a problem which {\it n}-th-hop of traversed entity should be taken. In fact, it would be quite time-consuming if we take more than two-hop entities into account. 

Given an extended route ({$_re_{m,1},_re_{m,2},_re_{m,3},...,_re_{m,k}$}) where $_re_{m,k}$ represents the \textit{k}-th hop extension in route \textit{r} of central entity \textit{m}, $_re_{m,k}$ has influence on $_re_{m,k-1}$, and $_re_{m,k-1}$ has influence on $_re_{m,k-2}$,..., $_re_{m,2}$ has influence on $_re_{m,1}$. We donate the influence as $\Leftarrow$ and elements a set, and the smaller index means more influence. So, we can get that,  $_re_{m,k-1}$ $\Leftarrow$ ($_re_{m,k}$), 
$_re_{m,k-2}$ $\Leftarrow$ ($_re_{m,k-1}$,$_re_{m,k}$),..., $_re_{m,1}$ $\Leftarrow$ ($_re_{m,2}$,$_re_{m,3}$,...,$_re_{m,k}$). After simple computing, we get that:

\vspace{-0.3cm}
\begin{small}
\begin{align}
& _re_{m,1} \Leftarrow (_re_{m,2},_re_{m,3},...,_re_{m,k-1},_re_{m,k})  \notag \\
& _re_{m,1} \Leftarrow ((_re_{m,3},...,_re_{m,k}),...,(_re_{m,k}), 0)  \label{con:mk1}
\end{align}
\end{small}
\vspace{-0.2cm}

In Equation \ref{con:mk1}, every former element in the set cover the information of the later ones, which satisfies the Markov property (sometimes characterized as ``memory lessness") \cite{spitzer1970interaction}. Therefore, we can convert the problem mentioned above into a Markov chain. In other words, we only take the first hop extensions into account while traversing entities in KGs of the central entities.

\begin{figure}[!ht]
    \centering
    \includegraphics[scale=0.5]{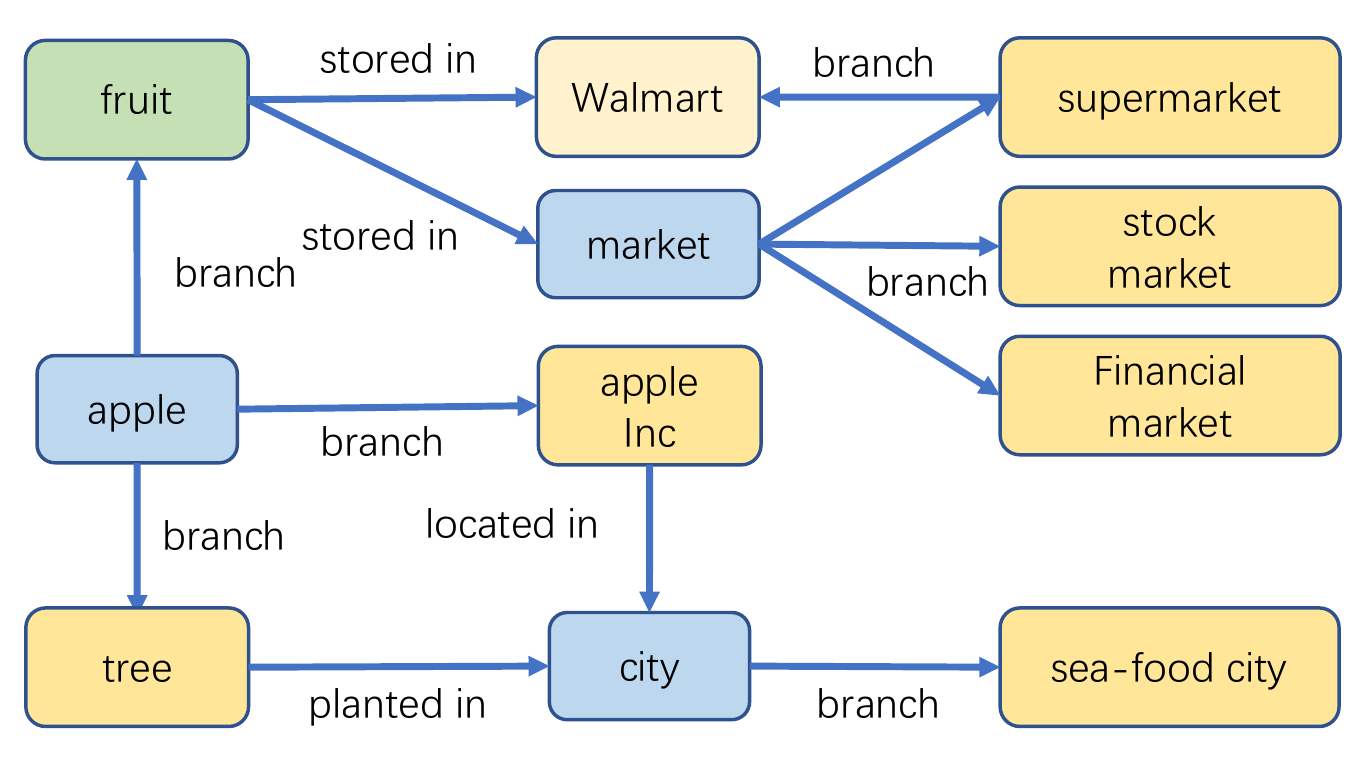}
    \caption{Partial Knowledge Graph from Dbpedia. The figure shows that `apple' could be a kind of food, as well as a company. We can get the precise meaning of `apple' by the neighbours around it.}
    \label{fig:relationship}
\end{figure}

To be specific and easy-understanding, we give an example to describe the process of extension. Given a sentence ``People in cities usually buy apples in the local markets.", we employ tool postag \cite{de2008stanford} of package nltk in python to extract properties of words in the sentence and we get that [(`People', `NNS'), (`in', `IN'), (`cities', `NNS'), (`usually', `RB'), (`buy', `VBP'), (`apples', `NNS'), (`in', `IN'), (`the', `DT'), (`local', `JJ'), (`markets', `NNS')]. CKG can easily capture the entities $[$`people', `city', `apple', `market'$]$ because they all are nouns. Then we traverse the central entities in knowledge graph and extend the first hop extensions of them which is described as Figure \ref{fig:relationship}. Every entity may have many extensions in KGs, but for the simplicity and low computing cost, we only choose the top 3 extensions of each central entity by the similarity calculation of their GloVe vectors. For the convenience, we donate the extensions of entity \textit{e} as set dt$[$`e'$]$. So, dt$[$`apple'$]$ = $[$`tree', `culture', `fruit'$]$. However, in general, each word in a specific sequence has only one meaning. Therefore, it turns to be a problem that which extension will be used to emphasize the meaning of the central entity. This will be discussed in Section \ref{sec:fusing}.

\vspace{-0.2cm}

\subsection{Fusing Extensions and Central Entities (Aggregator)} \label{sec:fusing}
After we extracted top 3 possible meanings for each central entities, it turns to be a problem that which one extension will be chosen to represent the true meaning of the central entity in a specific sequence. In this section, we present \textbf{Aggregator} to mitigate this problem.

In sequence \{$w_1,w_2,...,w_n$\}, we can get the initial vectors \{\bm{$w_1,w_2,...,w_n$}\} by the Equation \ref{con:glovew}. The extensions of entities in the sequence can get initial vectors by the same way and we get that \{\bm{$e_{1,1},...,e_{m,3}$}\}. CKG sums \{\bm{$w_1,w_2,...,w_n$}\} and then figure out the average of them, which is shown in Equation \ref{con:vavg}.

\vspace{-0.2cm}
\begin{small}
\begin{align}
& V_{avg}=\frac{\sum_{i} \bm{w_i}}{\{len(w_1,w_2,...,w_n)\}} \label{con:vavg} \\
& e_m \to \mathop{\arg\max}\limits_{k \epsilon \{e_{m,1},e_{m,2},e_{m,3}\}}(\cos{(GloVe(k), V_{avg})})  \label{con:ito}
\end{align}
\end{small}

Then, CKG figures out the cosine value between each of \{$e_{m,1},e_{m,2},e_{m,3}$\} with the average of the \{$w_1$,$w_2$,...,$w_n$\} and employ the one which the minimum angle to represent the meaning of entity \textit{m} in the sequence, which is shown in Equation \ref{con:ito}. In Equation \ref{con:ito}, we define a function $\cos{(a,b)}$ which is used to figure out the cosine value of vector \textit{a} and \textit{b}. Besides, we donate the relation ``with the meaning of " as symbol `$\to$', and  Function $\mathop{\arg\max}\limits_{k \epsilon \{e_{m,1},e_{m,2},e_{m,3}\}}(\cos{(GloVe(k), V_{avg})})$ means that we take the element \textit{k} which makes $\cos{(GloVe(k), V_{avg})}$ maximum.

In the example of Section \ref{sec:extension}, ${e_m}$= `apple', and top 3 extensions of apple in KGs is [`tree', `culture', `fruit'], which means dt$_{`apple'}$ = $[$`tree', `culture', `fruit'$]$ ({\it i.e., }for `apple', $e_1$ = `tree',$e_2$ = `culture', $e_3$ = `fruit'). After computing $e_i$ with $V_{avg}$ with Equation \ref{con:ito}, we can easily get that `apple' $\to$ `fruit'. Similar to `apple', `people' $\to$ `citizen', and `city' $\to$ `settlement', and `market' $\to$ `supermarket'.

\begin{figure}[!ht]
    \centering
    \includegraphics[scale=0.42]{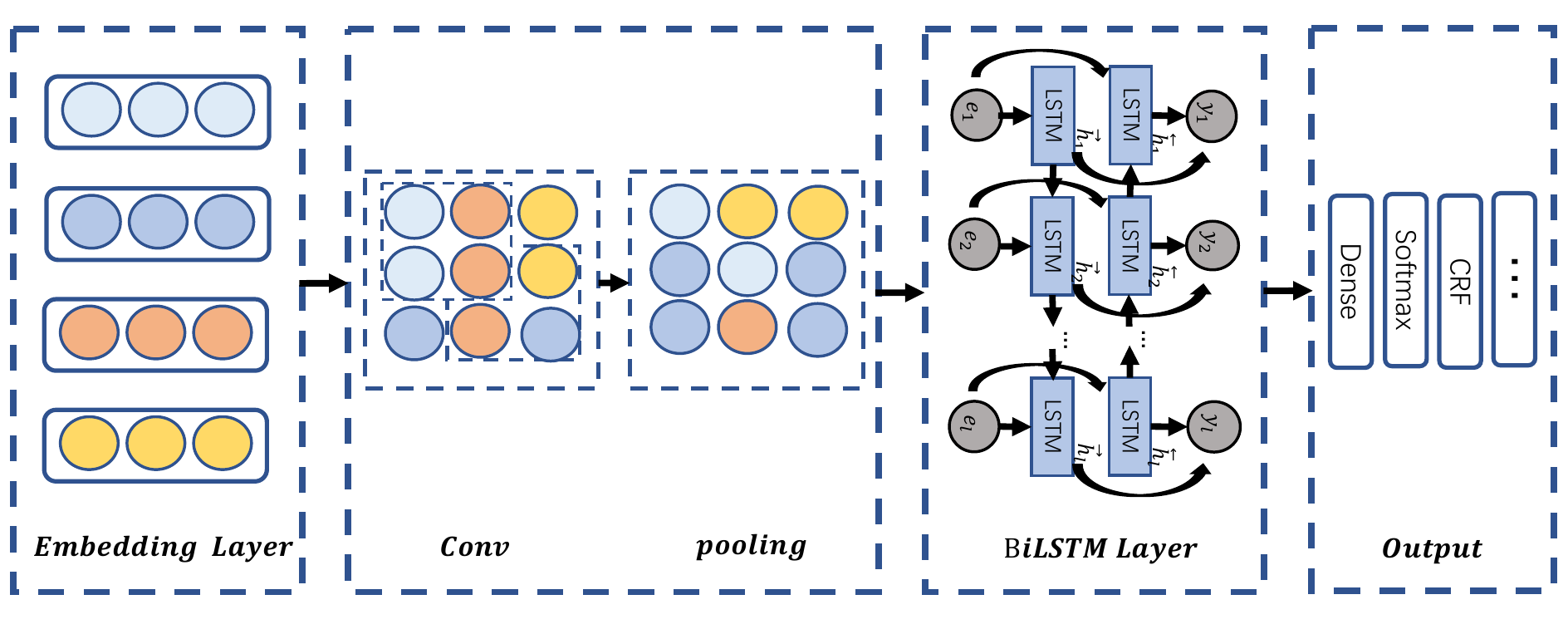}
    \caption{Semantic feature extraction from input embedding so that we can not only save the training time, but also avoid vanishing gradient.}
    \label{fig:model2}
\end{figure}

Finally, we add these precise meanings to the origin sequence with the form of note. As a result, we get a new sentence ``People (citizen) in cities (settlement) usually buy apples (fruit) in the local markets (supermarket)". And use this new sentence to replace the original input before training in Figure \ref{fig:model2} in the next section.

\subsection{Semantic Feature Extraction for Dimension Reduction} \label{sec:pre-training}
After semantic extension and fusing, we need to train a context-dependent representation model and test the performance of CKG. It proved that BiLSTM is a great sequence-to-sequence model which was widely used to train language models and has a good performance on most NLP benchmarks. Therefore, in this section, we take bidirectional LSTM to train our language model. 

However, BiLSTM has a step size limit to the length of input sequences because it is a sequential model. The general step size limit ranges from 250 to 500. If the length of the sequential step of BiLSTM is oversize, the gradient will vanish soon. Moreover, it is time-consuming for BiLSTM because the outputs of prior cells will be used as inputs of the next cell. While if we reduce the dimension of the input sequence of BiLSTM or the input is full features, the time costs for training would be greatly saved and the gradient vanishing would also be avoided.

\vspace{-0.2cm}

Before BiLSTM layer, we introduce a 1D-CNN layer for identifying patterns within data. The patterns will then be used to form more complex patterns in higher layers. So, after 1D-CNN extraction processing, the information of origin sequence ($t_1, t_2, \dots, t_n$) can be represented as a middle state embedding sequential ($c_1, c_2, \dots, c_m$) where \textit{m} $<$ \textit{n}. Then we put the middle sequential into BiLSTM layer in CKG. A forward language model computes the probability of the sequence by modeling the probability of $c_k$ given the history ($c_1, c_2, \dots c_{k-1}$).Each position k of forward LSTM will output a context-dependent representation $\overrightarrow{\mathbf{h}}_{k, j}^{LM}$ (where \textit{j} = 1, 2, $\dots$, \textit{L}). Similar to forward LSTM, the backward LSTM will get $\overleftarrow{\mathbf{h}}_{k, j}^{LM}$. We connect results of forward and backward LSTMs as ${\mathbf{h}}_{k, j}^{LM}$ = $\overrightarrow{\mathbf{h}}_{k, j}^{LM}$ + $\overleftarrow{\mathbf{h}}_{k, j}^{LM}$. The probability of the sequence is shown as Equation \ref{con: probpost}.

\vspace{-0.3cm}
\begin{small}
\begin{align}
& p(c_{1}, c_{2}, \ldots, c_{m})=\prod_{k=1}^{m} p(c_{k} | c_{1}, c_{2}, \ldots, c_{k-1}) \label{con: probpre} \\
& p(c_{1}, c_{2}, \ldots, c_{m})=\prod_{k=1}^{m} p(c_{k} | c_{k+1}, c_{k+2}, \ldots, c_{m})  \label{con: probpost}
\end{align}
\end{small}

\vspace{-0.3cm}

We design a loss function to adjust CKG parameters, which is shown as the Equation \ref{con: lossfuc}. The goal in our work is to make the Equation \ref{con: lossfuc} maximum. It needs to modify the weights and values of our network until we make the result of Equation \ref{con: lossfuc} maximize. And then we use output layer like Softmax layer, crf layer to do the downstream tasks.

\vspace{-0.3cm}

\begin{small}
\begin{equation}
    \log p(c_k | c_1, \dots, c_{k-1}) + \log p(c_{k} | c_{k+1}, \ldots, c_{m}))
\label{con: lossfuc}
\end{equation}
\end{small}

\vspace{-0.3cm}
\section{Experiments} \label{sec:experiment}
In this section, we present the details of pre-training CKG and the fine-tuning results on nine NLP datasets, which contain knowledge-driven tasks and common NLP tasks. In this paper, we train BiLSTM on a corpus with approximately 30 million sentences \cite{chelba2013one} which is mentioned in \cite{Peters:2018}. The pre-trained biLSTM here is similar to the architectures in \cite{Peters:2018}.

\subsection{Qualitative Argument} \label{sec:qa}
In order to demonstrate the performance of enhancing semantic in specific context, we conduct CKG on an entity typing task with a well-established dataset FIGER \cite{ling-etal-2015-design}. The training set of FIGER is labeled with distant supervision, and its test set is annotated by human. Following the evaluation method in \cite{shimaoka2016attentive}, we compare NFGEC, ERNIE, CKG on FIGER, and adopt strict accuracy, loose macro, loose micro scores for evaluation standard. The experimental results on FIGER is shown in Table \ref{tab:figer}, the baseline methods for entity typing we compare our models with are as follows:

\begin{itemize}
    \item \textbf{NFGEC} \cite{shimaoka2016attentive}. NFGEC is the SOAT model on FIGER, which combines the representations of three-granularity ({i.e., entity mention, context, and extra hand-craft}) features as input.
    \item \textbf{ERNIE} \cite{zhangetal2019ernie}. ERNIE is a pre-trained method considering both large-scale textual corpora and KGs, which outperforms on FIGER.
\end{itemize}

\begin{table}[ht]
\caption{Results of various models on FIGER(\%)} 
\centering
\resizebox{0.47\textwidth}{!}{
\begin{tabular}{c|c|c|c|c}
\hline\hline
Model & \begin{tabular}[c]{@{}c@{}}NFGEC\\ (Attentive)\end{tabular} & \begin{tabular}[c]{@{}c@{}}NFGEC\\ (LSTM)\end{tabular} & \begin{tabular}[c]{@{}c@{}}ERNIE\\ (tsinghua)\end{tabular} & CKG \\ \hline
Acc. & 54.53 & 55.60 & 57.19 & 58.84 \\ \hline
Macro & 74.76 & 75.15 & 76.51 & 76.23 \\ \hline
Micro & 71.58 & 71.73 & 73.39 & 75.24 \\ \hline\hline
\end{tabular}}
\label{tab:figer}
\end{table}

Table \ref{tab:figer} shows the results in FIGER, and we can find that: (1) ERNIE achieves higer accuracy than NFGEC method, indicating the external knowledge regularizes ERNIE to avoid fitting the noisy labels and accordingly benefits entity typing. (2) CKG has lower macro than the best ERNIE method, but it significantly improves the strict accuracy and micro, which achieves 58.87\% in accuracy (2.89\% higher than ERNIE) and 76.23\% (2.52\% higher than the second score). Obviously, CKG can be used to reduce the noisy label challenge in FIGER by injecting the external information from KGs.

\subsection{Benchmarks} \label{sec:benchmark}
To make a objectively comparison with baseline models, we conduct experiments separately on the following datasets.

\begin{itemize}
	\item Word analogy task \cite{mikolov2013distributed}. The dataset is divided into 14 subsets and made up of 19,544 question like ``a is to b like c is to d''.
	\item Quora Question Pairs (QQP) \cite{chen2018quora}. QQP is a binary classification task including 400,000 question pairs, where we need to value if two questions in each pair are equal or not.
	\item Microsoft Research Paraphrase Corpus (MRPC) \cite{dolan2005automatically}. MRPC contains sentence pairs which are extracted from online news sources, in which there are annotations for whether the sentences are equal in semantic from human.
	\item Stanford Sentiment Treebank (SST-5) \cite{socher2013recursive}. SST-5 is a binary single-sentence classification task from the Stanford Sentiment Tree bank, including five levels of comment(from very negative to very positive), used to describe a sentence out of a movie review.
	\item Stanford Question Answering Dataset (SQuAD v1.1) \cite{rajpurkar2016squad}.SQuAD consists of questions raised by a crowd of workers on a set of Wikipedia articles, each of which is a piece of text or span from the corresponding reading paragraph.Each pair contains a question and a passage where we can find the answer.
	\item Named Entity Extraction (NER) \cite{sang2003introduction}. NER includes 1,393 English and 909 German news articles. To build the English-language corpus we need the RCV1 Reuters corpus where entities are annotated with LOC (location), ORG (organisation), PER (person) and MISC (miscellaneous).
	\item Stanford Natural Language Inference (SNLI) \cite{bowman2015large}. SNLI takes a pair of sentences and predicts if the former one cover the meaning of the later one, which collects 570k English sentence pairs written by human and labeled with entailment, contradiction and neutral as the judgment. 
\end{itemize}

\subsection{Results}

\begin{table}[!ht]
\caption{Comparison among CKG+ELMo and other models in QQP, SST-5, and MRPC tasks}
\centering
\begin{tabular}{c|c|c|c}
\hline\hline
Baseline & \begin{tabular}[c]{@{}c@{}}Pre-OpenAI\\ SOTA\end{tabular} & \begin{tabular}[c]{@{}c@{}}BiLSTM+\\ ELMo+Attn\end{tabular} & \begin{tabular}[c]{@{}c@{}}ERNIE\\ (tsinghua)\end{tabular} \\ \hline
QQP & 93.2 & 90.4 & 93.5 \\ \hline
MRPC & 86.0 & 84.9 & 88.9 \\ \hline
SST-5 & 66.1 & 64.8 & 71.2 \\ \hline\hline
 & \begin{tabular}[c]{@{}c@{}}ERNIE2.0\\ (baidu)\end{tabular} & CKG & CKG+ELMo \\ \hline
QQP & 90.4 & 92.7 & 93.2 \\ \hline
MRPC & 88.9 & 88.2 & 86.3 \\ \hline
SST-5 & 70.4 & 70.2 & 72.3 \\ \hline\hline
\end{tabular}
\label{tab:qt}
\end{table}

\textbf{Word analogy task}. We combine the part one of CKG shown as Figure \ref{fig:model1} with GloVe to retrain wikipedia2014 and get a new word vectors which improve the score on word analogy task to 79.37. Main details are shown in the Table \ref{tab:wordsim_analogy}. Table \ref{tab:wordsim_analogy} indicates that we get a higher score based spearman algorithm, comparing with different models like CBOW, SG, and pure Glove. Table \ref{tab:wordsim_analogy} shows CKG+GloVe gets 10.44 score more than pure GloVe on word analogy task.

\begin{table}[!ht]
\caption{Comparison of models in word similarity with rank of sepearman. And the  comparison of word analogy in semantic, syntactic, average.}
\centering
\resizebox{0.47\textwidth}{!}{
\begin{tabular}{c|c|c|c|c}
\hline
\hline
Baseline & Semantic & Syntactic & Average & Rank of Spearman \\ \hline
CBOW & 73.58 & 65.95 & 69.5 & 73.25 \\ \hline
SG & 65.62 & 56.61 & 60.64 & 68.69 \\ \hline
GloVe & 71.39 & 53.72 & 61.57 & 68.93 \\ \hline
CKG+GloVe & 78.34 & 69.32 & 73.83 & 79.37 \\ \hline\hline
\end{tabular}}
\label{tab:wordsim_analogy}
\end{table}

\begin{figure}[!ht]
    \centering
    \includegraphics[scale=0.37]{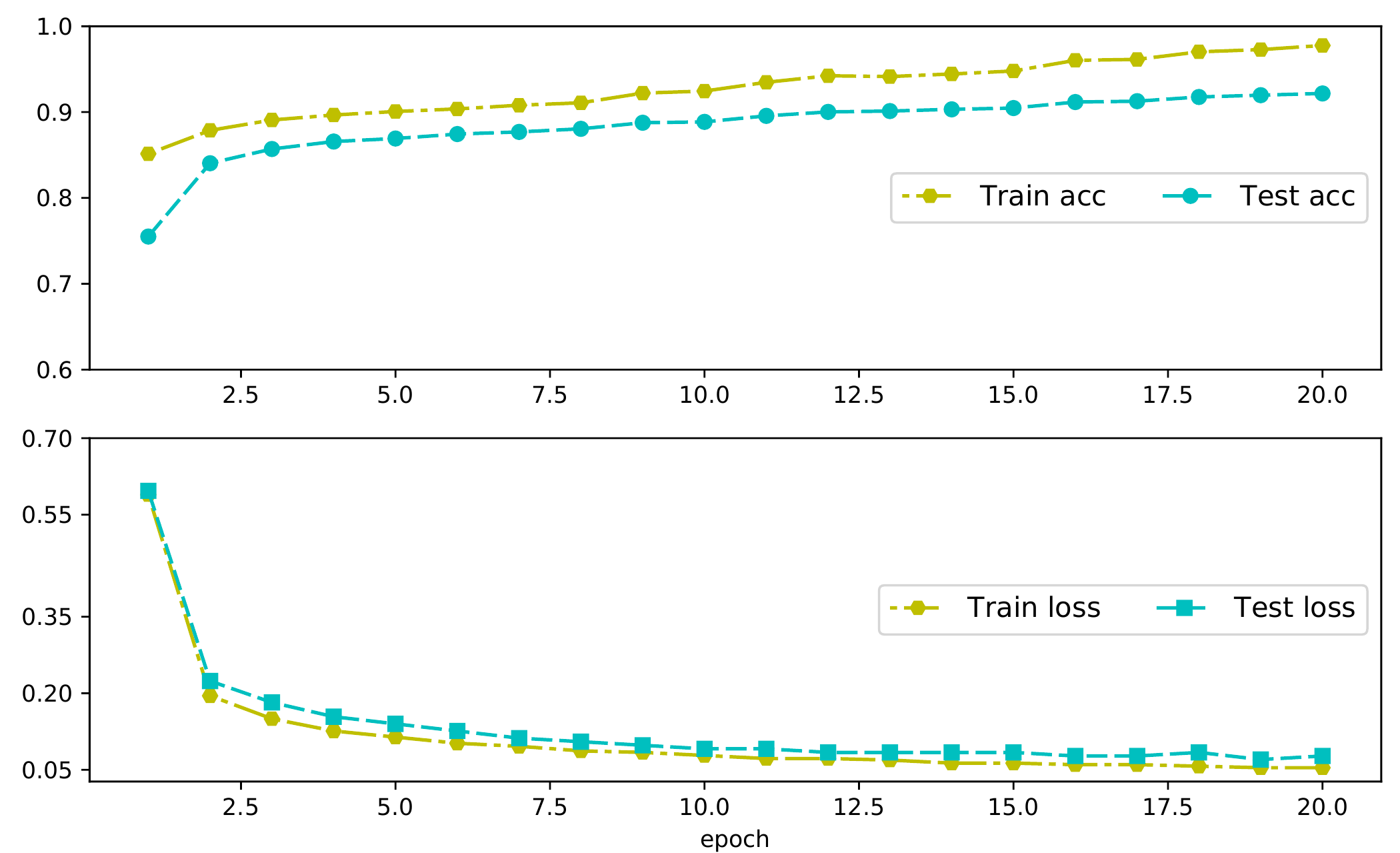}
    \caption{The detailed information on the dataset of SNLI. The upper figure represents the accuracy results of training and test in different epochs, and the lower figure indicates the losses of them.}
    \label{fig:snli}
\end{figure}

\vspace{0.2cm}
\textbf{QQP, MRPC, and SST-5}. In QQP, CKG+ELMo gets score 93.2 equals Pre-OpenAI SOTA and better than pure ELMo (90.4) (shown as Table \ref{tab:qt}), but ERNIE performs better than our CKG. In both MRPC and SST-5, CKG+ELMo gets score 86.3/72.3 which is better than the other Pre-OpenAI SOTA and pure ELMo. In addition, CKG+ELMo performs better than other models on SST-5.

\vspace{0.2cm}
\textbf{SQuAD (v1.1)}. Our baseline \cite{Peters:2018} is an improved version of an unsupervised LM in \cite{peters2017semi}. It is deep, and in the sense, ELMo representations are a function dealing with all of the internal layers of the biLM. We combine CKG and BiDAF \cite{seo2016bidirectional} to conduct experiment on SQuAD. Table \ref{tab:SQuAD} shows results on the \cite{rajpurkar2016squad} with combination of CKG and ELMo. In \cite{rajpurkar2016squad}, there will give a question and a paragraph containing the answer to the former question. ELMo gets score 85.8 based on the baseline of SAN with BiLSTM. $BERT_{BASE}$ gets 88.5 with its outstanding pre-training parameters and fine-tuning part. Above all, CKG+ELMo get a higher score of 89.2, which proves the dynamic adjustment model in CKG effective in natural language processing task. As a fine-tuning tool, the CKG always performs well considering the global factor.

\begin{table}[!ht]
\caption{Analysis for SQuAD, SNLI, and NER, comparing different choices. We compares systems with pure GloVe, ELMo, the combination of GloVe and CKG shown as Figure \ref{fig:model1} and only CKG. The set of comparison of the three models. ELMo has achieved a great performance in NER task, while CKG performs better.} 
\centering
\resizebox{0.45\textwidth}{!}{
\begin{tabular}{c|c|c|c|l}
\hline\hline
Task & GloVe & ELMo & CKG+GloVe & CKG \\ \hline
SQuAD & 80.8 & 85.8 & 85.6 & 88.7 \\ \hline
SNLI & 88.1 & 89.1 & 90.2 & 91.1 \\ \hline
NER & 87.7 & 91.9 & - & 92.56 \\ \hline\hline
\end{tabular}}
\label{tab:sqsnli}
\end{table}

\begin{table}[!ht]
\caption{The set of comparison of the five models. We compare SAN, ELMo, BERT$_{BASE}$, with CKG and CKG+ELMo across SQuAD task. The “INCREASE” column lists improvement over our baseline.}
\centering
\resizebox{0.47\textwidth}{!}{
\begin{tabular}{c|c|c|c|c|c}
\hline\hline
Model & SAN & ELMo & \begin{tabular}[c]{@{}c@{}}BERT\\ (base)\end{tabular} & CKG & \begin{tabular}[c]{@{}c@{}}CKG\\ + ELMo\end{tabular} \\ \hline
SOTA & 84.4 & 85.8 & 88.5 & 88.7 & 89.2 \\ \hline
INCREASE & baseline & 1.66\% & 4.86\% & 5.09\% & 5.69\% \\ \hline\hline
\end{tabular}}
\label{tab:SQuAD}
\end{table}

\textbf{SNLI}. Our baseline is \cite{Peters:2018}, and we combine CKG with ESIM \cite{chen2017enhanced} to deal with semantic entailment task. ESIM is a sequence model which uses a biLSTM to encode the premise and hypothesis. And ESIM has three layers including a matrix attention layer, a local inference layer, and biLSTM inference composition layer, followed by a pooling operation before the output layer. In general, the combination of CKG and ESIM improves accuracy by an average of 2.7\%. CKG pushes the overall accuracy to 91.1\%, exceeding \cite{Peters:2018}.

\vspace{0.2cm}
\textbf{NER}. Our baseline \cite{peters2017semi,Peters:2018} uses pre-trained word embeddings trained on a big corpus, and includes a character-based CNN representation for word-level scalability, two biLSTM layers for hidden states and a conditional random field (CRF) loss. As shown in Table \ref{tab:sqsnli}, CKG with CRF achieves the score 92.56. The main difference between CKG and the previous great work from \cite{Peters:2018} is that we took the semantic refining before pre-training into consideration. 

\section{Conclusion}
This paper presents a new model called CKG, including extending and refusing semantic of entities by KGs. To obtain more meanings of a polysemous entity, we traverse its neighbours in Dbpedia, and get neighbours within one-hop as its extensions. Then we choose one most similar to the average meaning of the rest entities to represent the true meaning of the central entity. As a result, we can mitigate polysemy problem in NLP. In addition, CKG can improve the semantic in entity type classification, and outperforms in multiple NLP tasks.

\bibliographystyle{IEEEtran}
\bibliography{IEEEabrv,IEEEexample}
\end{document}